\documentclass{article}

\usepackage{microtype}
\usepackage{graphicx}
\usepackage{subcaption}
\usepackage{booktabs}
\usepackage[breaklinks=true]{hyperref}

\usepackage[accepted]{icml2026}

\usepackage{amsmath}
\usepackage{amssymb}
\usepackage{mathtools}
\usepackage{amsthm}

\usepackage[capitalize,noabbrev]{cleveref}

\theoremstyle{plain}

\theoremstyle{definition}

\theoremstyle{remark}

\icmltitlerunning{Sparse Memory Finetuning as a Low-Forgetting Alternative to LoRA and Full Finetuning}

\begin{document}

\twocolumn[
  \icmltitle{Sparse Memory Finetuning as a Low-Forgetting\\
    Alternative to LoRA and Full Finetuning}

  \icmlsetsymbol{equal}{*}

  \begin{icmlauthorlist}
    \icmlauthor{Prakhar Gupta}{equal,umich}
    \icmlauthor{Garv Shah}{equal,umich}
    \icmlauthor{Satyam Goyal}{equal,umich}
    \icmlauthor{Anirudh Kanchi}{equal,umich}
  \end{icmlauthorlist}

  \icmlaffiliation{umich}{University of Michigan}

  \icmlcorrespondingauthor{Prakhar Gupta}{prakharg@umich.edu}

  \icmlkeywords{Continual Learning, Memory Layers, Parameter-Efficient Finetuning, Catastrophic Forgetting}

  \vskip 0.3in
]

\printAffiliationsAndNotice{\icmlEqualContribution}

\begin{abstract}
Adapting a pretrained language model to a new task often hurts the
general capabilities it already had, a problem known as catastrophic
forgetting. Sparse Memory Finetuning (SMF) tries to avoid this by
adding key--value memory layers to the model and, on each training
step, updating only the small set of memory rows that the current
batch reads most heavily. We re-implement SMF on
\texttt{Qwen-2.5-0.5B-Instruct} and compare it with LoRA and full
finetuning on MedMCQA, a 4-choice medical exam task, using WikiText
perplexity and TriviaQA accuracy as forgetting probes. SMF improves
MedMCQA by $2.5$ percentage points while keeping both forgetting
probes within roughly $1$ point of the base model, whereas LoRA and
full finetuning achieve larger gains but with clear drift on both. We
also compare two row-selection rules (KL-divergence and TF-IDF), which
balance the two forgetting metrics differently.
\end{abstract}

\section{Introduction}
\label{sec:intro}

Adapting a pretrained language model to a narrow domain typically uses
either parameter-efficient methods (such as LoRA \citep{hu2022lora}) or
full finetuning. Both approaches modify weights that affect every input,
so adapting to a new task can degrade performance on capabilities the
model already had, an effect often called catastrophic forgetting.

Sparse Memory Finetuning (SMF)\footnote{Code: \url{https://github.com/prakharg55/SMF-ICML-FG}} \citep{lin2025sparse} takes a different
approach. It first inserts key--value memory layers
\citep{lample2019large,berges2024memory} into selected transformer
layers, then on each step trains only the small subset of memory value
rows that the current batch reads most heavily, leaving the rest
frozen.

\newpage

In this work we re-implement SMF on
\texttt{Qwen-2.5-0.5B-Instruct} \citep{qwen2024} and run a controlled
comparison against LoRA and full finetuning. We evaluate task learning
on MedMCQA \citep{pal2022medmcqa}, a 4-choice medical exam dataset, and
measure forgetting on WikiText perplexity \citep{merity2017pointer}
(general language modeling) and TriviaQA accuracy
\citep{joshi2017triviaqa} (open-domain factual recall).

Our contributions are:
\begin{itemize}
  \item A controlled comparison of sparse memory finetuning, LoRA, and
        full finetuning on MedMCQA across random seeds, evaluated
        jointly on a target-task metric and two distinct
        forgetting probes.
  \item A side-by-side comparison of the TF-IDF slot-selection rule of
        \citet{lin2025sparse} and a novel KL-divergence rule within each
        sparse architecture.
\end{itemize}

We find that additive sparse memory finetuning with KL slot selection
occupies a distinct point on the plasticity--stability frontier:
smaller MedMCQA gain than LoRA or full finetuning ($+2.5$ pp vs
$+4.6$ and $+5.4$ pp), but with substantially less drift on both
forgetting probes.

\section{Related Work}
\label{sec:related}

LoRA \citep{hu2022lora} adapts a pretrained model by adding low-rank
trainable updates to existing linear projections, and is widely used
because it is simple and often close to full finetuning in target-task
quality. Product Key Memory \citep{lample2019large} adds a large
key--value table addressed by factorized product keys; later
memory-layer variants scale and modify it for modern transformers
\citep{berges2024memory}. SMF \citep{lin2025sparse} uses these memory
layers as an adaptation substrate rather than as a pretraining-time
capacity expansion. Our comparison asks whether such a localized
memory update can trade some target-task plasticity for less drift on
unrelated capabilities. In the language-model setting this is hard
to summarize with one number, so we evaluate both general language
modeling (WikiText perplexity) and open-domain factual recall
(TriviaQA).

\section{Method}
\label{sec:method}

\begin{figure*}[t]
  \centering
  \includegraphics[width=0.95\textwidth]{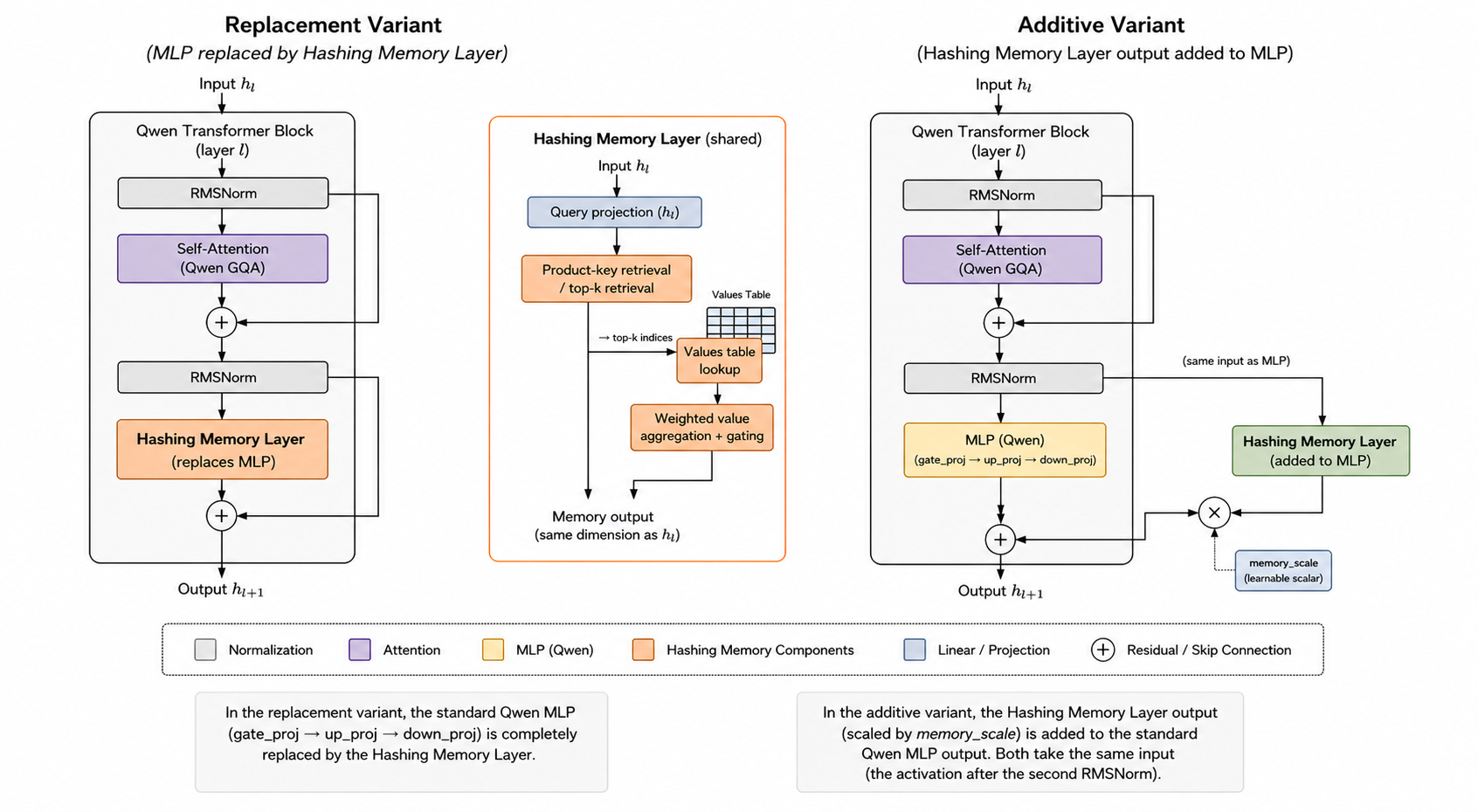}
  \caption{Method overview. Each Qwen-2.5 transformer block uses
  RMSNorm, grouped-query self-attention, and a SwiGLU MLP
  $\mathrm{down\_proj}(\mathrm{silu}(\mathrm{gate\_proj}(x)) \odot
  \mathrm{up\_proj}(x))$. We compare two ways of inserting a Hashing
  Memory Layer at selected layers: \textbf{Replacement} substitutes the
  MLP entirely (left); \textbf{Additive} keeps the MLP and adds a
  memory-scaled branch (right). The middle panel details the memory
  layer (Section~\ref{sec:memory}): queries are matched against
  product-key tables, the top-$k$ value rows are softmax-weighted, and
  the result passes through a SiLU-gated, SwiGLU-style output projection.}
  \label{fig:method}
\end{figure*}

\subsection{Memory Layer Architecture}
\label{sec:memory}

We use the Product Key Memory (PKM) architecture of
\citet{lample2019large} in the scaled-up form proposed by
\citet{berges2024memory}, which is the same family of memory layers used
by \citet{lin2025sparse}. PKM stores $M = n_k^2$ value rows
$V \in \mathbb{R}^{M \times d}$
and a set of trainable keys factored as the outer product of two
sub-key tables. For an input $h$ a query $q = W_q h$ is split in half,
matched against each sub-key table, and the top-$k$ overall slots
$\mathcal{I}$ are retrieved. The retrieved values are combined into a
softmax-weighted sum
\begin{equation}
  r(h) = \sum_{i \in \mathcal{I}} p_i \, v_i,
  \quad p_i = \mathrm{softmax}(q \cdot k_i)_{i \in \mathcal{I}},
\end{equation}
which is then passed through the SiLU-gated, SwiGLU-style value projection
$\mathrm{mem}(h) = W_o \big( r(h) \odot \mathrm{silu}(W_g h) \big)$,
following \citet{berges2024memory}. We insert separate PKM modules at
layers $\{6, 12, 18\}$ of Qwen with $n_k = 128$ (so $M = 16{,}384$
slots per layer), 4 heads, $k = 16$ top neighbors per token, and key
dimension 256. Memory value tables are not shared across layers.

We compare two ways of integrating the memory at each chosen layer
(Figure~\ref{fig:method}): \textbf{Replacement}
($\mathrm{MLP}_\ell \leftarrow \mathrm{mem}_\ell$)
discards the original Qwen MLP, and \textbf{Additive}
($\mathrm{MLP}_\ell(h) + \alpha \cdot \mathrm{mem}_\ell(h)$, with
$\alpha$ initialized to $0.01$) keeps the original MLP and adds a
scaled memory branch. We also study an \textbf{Additive +S} variant
in which the per-layer scalar $\alpha$ is trainable during sparse task
training.

\subsection{Sparse Update via Top-$T$ Selection}
\label{sec:sparse}

For each training batch we count how many times each value row is read
by the forward pass. Let $c(i)$ denote the read count for slot $i$ in
the current batch (summed across heads, tokens, and top-$k$ neighbors)
and $C = \sum_j c(j)$. We score each accessed slot, mask the others
with $-\infty$, and select the top-$T$ slots per layer. A backward hook
on the value matrix multiplies the gradient by a boolean mask so only
selected rows accumulate gradient. All other rows are frozen for that
step. We use $T = 512$ throughout.

We compare two scoring rules. The \textbf{TF-IDF} rule of
\citet{lin2025sparse} uses a background corpus statistic: let $df(i)$
be the number of background batches in which slot $i$ was read at
least once, and $N$ the total number of background batches. Then
\begin{equation}
  s_{\mathrm{tfidf}}(i) = \frac{c(i)}{C} \cdot \log\!\left(\frac{N + 1}{df(i) + 1}\right).
\end{equation}
This rewards slots that are read often in the current batch but rarely
in background.

The \textbf{KL} rule we use compares two distributions of the same
type: let $p_{\mathrm{batch}}(i) = c(i)/C$ and $p_{\mathrm{bg}}(i) =
(b(i) + 1) / \sum_j (b(j) + 1)$, where $b(i)$ is the cumulative
\emph{token-level} read count of slot $i$ across all background batches
(i.e., the same kind of quantity as $c(i)$). The score is the
per-slot contribution to $D_{\mathrm{KL}}(p_{\mathrm{batch}} \,\Vert\,
p_{\mathrm{bg}})$:
\begin{equation}
  s_{\mathrm{kl}}(i) = p_{\mathrm{batch}}(i) \cdot \log\!\left(\frac{p_{\mathrm{batch}}(i) + \varepsilon}{p_{\mathrm{bg}}(i) + \varepsilon}\right).
\end{equation}
Both rules favor slots that are over-used in the current batch
relative to background. They differ in whether the background
reference is a document-frequency or a token-frequency distribution.

\subsection{Two-Stage Training Pipeline}
\label{sec:pipeline}

\textbf{Stage 1: dense retrofit.} The original Qwen MLP weights at
layers $\{6, 12, 18\}$ are either replaced or augmented with a freshly
initialized memory layer (Section~\ref{sec:memory}). All non-memory
parameters are frozen and the memory parameters are trained for 2
epochs on 50{,}000 OpenAssistant \citep{kopf2023openassistant}
assistant responses. This stage gives the memory layers a sane
initialization before sparse task training.

\textbf{Stage 2: sparse task training.} Starting from the retrofit
checkpoint, only the value rows selected by the per-batch top-$T$
mask receive gradient. The base Qwen parameters and the memory keys
remain frozen (for Additive +S, the scalar $\alpha$ is also trainable).
We train for 3 epochs on 60{,}000 MedMCQA training examples.

We compare against two non-sparse baselines. The first is
\textbf{LoRA} \citep{hu2022lora} with rank 16, $\alpha_{\mathrm{LoRA}}
= 32$, dropout 0.05, applied to all attention and MLP linear
projections. The second is \textbf{full finetuning} of all Qwen
parameters. Both baselines train on the same 60{,}000 MedMCQA examples
for 3 epochs. During Stage 2, at most $3{\times}512 \approx 1.38$M
memory value parameters receive gradient per step, versus ${\sim}9$M
for LoRA and ${\sim}494$M for full finetuning; full storage,
inference-size, and per-step update accounting is in
Appendix~\ref{app:overhead}.

\section{Experiments}
\label{sec:exp}

\subsection{Setup}
\label{sec:setup}

We use \texttt{Qwen-2.5-0.5B-Instruct} \citep{qwen2024} as the base
model. Sparse and LoRA training use a learning rate of $5\!\times\!10^{-4}$
and $2\!\times\!10^{-4}$ respectively. Full finetuning uses
$5\!\times\!10^{-5}$. All training runs use a global batch size of 16
and the AdamW optimizer with cosine learning-rate schedule.

We evaluate three metrics on a held-out 1{,}000-example slice each:
MedMCQA accuracy as the target task, WikiText-103
\citep{merity2017pointer} test perplexity as a language-modeling
forgetting probe, and TriviaQA \citep{joshi2017triviaqa} (rc.nocontext
validation) accuracy via alias-substring match as a knowledge
forgetting probe. MedMCQA is scored by mean per-token log-likelihood of
each \texttt{"\{label\}. \{option\}"} continuation given the question
prompt. WikiText perplexity uses standard sliding-window evaluation
with window size 1024 and stride 512.

Background statistics for sparse slot selection are collected from
2{,}000 single-example batches of OpenAssistant.

\subsection{Main Results}
\label{sec:results}

\begin{figure*}[t]
  \centering
  \includegraphics[width=0.95\textwidth]{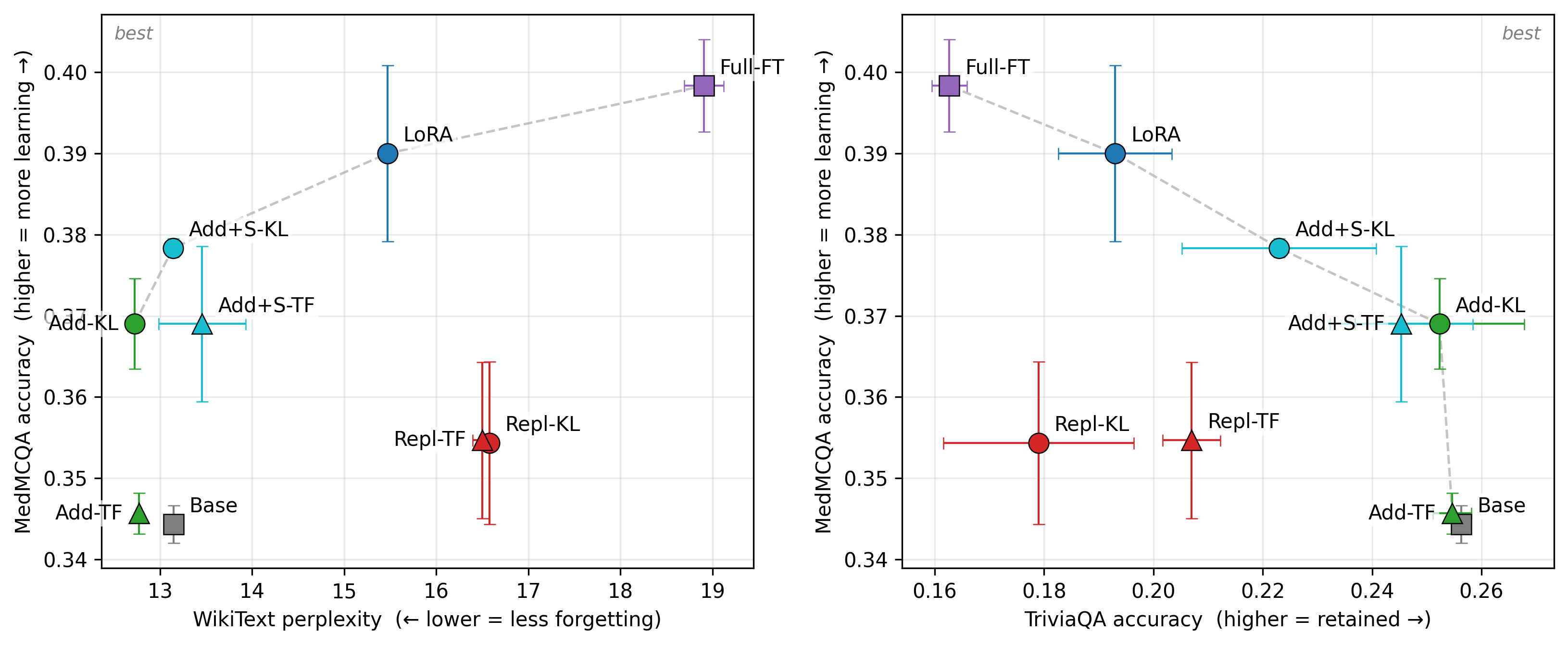}
  \caption{Plasticity--stability frontier on MedMCQA. Each method
  appears as one point with cross-seed standard-deviation error bars.
  Color encodes the method family. For sparse methods, marker shape
  distinguishes the slot-selection rule (circle $=$ KL, triangle $=$
  TF-IDF). Non-sparse baselines (Base Qwen, LoRA, Full finetune)
  appear as their own labeled points. Top-left of the left panel and
  top-right of the right panel are best. Dashed lines mark the
  empirical Pareto frontier.}
  \label{fig:pareto}
\end{figure*}

Table~\ref{tab:results} reports mean $\pm$ standard deviation across
random seeds for the nine evaluated conditions. Figure~\ref{fig:pareto}
shows the same data as a Pareto scatter on each forgetting axis.

\begin{table*}[t]
  \caption{MedMCQA accuracy ($\uparrow$), WikiText test perplexity
  ($\downarrow$), and TriviaQA alias-substring accuracy ($\uparrow$) on
  a 1{,}000-example evaluation slice. Mean $\pm$ standard deviation
  across random seeds. Bold marks the best in each column.}
  \label{tab:results}
  \vskip 0.05in
  \begin{center}
    \begin{small}
      \begin{tabular}{lccc}
        \toprule
        Method & MedMCQA acc $\uparrow$ & WikiText PPL $\downarrow$ & TriviaQA acc $\uparrow$ \\
        \midrule
        Base Qwen                        & 0.344 $\pm$ 0.002 & 13.146 $\pm$ 0.000          & \textbf{0.256} $\pm$ 0.001 \\
        Replacement sparse (KL)          & 0.354 $\pm$ 0.010 & 16.577 $\pm$ 0.045          & 0.179 $\pm$ 0.017 \\
        Replacement sparse (TF-IDF)      & 0.355 $\pm$ 0.010 & 16.501 $\pm$ 0.106          & 0.207 $\pm$ 0.005 \\
        Additive sparse (KL)             & 0.369 $\pm$ 0.006 & \textbf{12.723} $\pm$ 0.039 & 0.252 $\pm$ 0.016 \\
        Additive sparse (TF-IDF)         & 0.346 $\pm$ 0.003 & 12.773 $\pm$ 0.025          & 0.255 $\pm$ 0.004 \\
        Additive sparse +S (KL)          & 0.378 $\pm$ 0.001 & 13.139 $\pm$ 0.068          & 0.223 $\pm$ 0.018 \\
        Additive sparse +S (TF-IDF)      & 0.369 $\pm$ 0.010 & 13.456 $\pm$ 0.472          & 0.245 $\pm$ 0.013 \\
        LoRA                             & 0.390 $\pm$ 0.011 & 15.470 $\pm$ 0.072          & 0.193 $\pm$ 0.010 \\
        Full finetune                    & \textbf{0.398} $\pm$ 0.006 & 18.907 $\pm$ 0.215 & 0.163 $\pm$ 0.003 \\
        \bottomrule
      \end{tabular}
    \end{small}
  \end{center}
\end{table*}

\textbf{Replacement memory is dominated.} Both replacement-sparse
variants improve MedMCQA only marginally (+1.0 to +1.1 pp over base)
while increasing WikiText perplexity by about 26\% and reducing
TriviaQA accuracy by 5 to 8 pp. They lie outside both Pareto frontiers
in Figure~\ref{fig:pareto}. This matches the conclusion of
\citet{lin2025sparse} that overwriting pretrained MLPs is too
destructive at this scale.

\textbf{Additive sparse memory preserves both forgetting metrics
without sacrificing task gain.} Additive sparse (KL) gains $+2.5$ pp on
MedMCQA (0.344 to 0.369) while WikiText perplexity moves from $13.15$
to $12.72$, a small absolute change that is in fact slightly below
base, and TriviaQA falls only $0.4$ pp ($0.256 \to 0.252$, within the
trained-method seed spread). Among the methods that meaningfully
improve MedMCQA, this is the clearest low-forgetting point we observe:
neither forgetting metric degrades by more than $1$ pp (or $1\%$)
relative to base. The
Additive sparse +S (TF-IDF) variant is comparable: $+2.5$ pp on
MedMCQA, $-1.1$ pp on TriviaQA, and a $+0.31$ WikiText perplexity
drift that is within the trained-method seed spread.

\textbf{LoRA and full finetuning learn more but forget more.} Full
finetuning achieves the best MedMCQA accuracy ($0.398$) but raises
WikiText perplexity by $44\%$ and reduces TriviaQA accuracy by $9.3$
pp, the largest forgetting on either axis. LoRA achieves a slightly
smaller MedMCQA gain ($0.390$) with milder forgetting (a $17\%$
WikiText perplexity increase and a $6$ pp TriviaQA drop). In
Figure~\ref{fig:pareto} both methods sit on the Pareto frontiers but
at the high-forgetting end of each, far from the additive sparse
variants which trade $1$--$2$ pp of MedMCQA accuracy for substantially
less drift.

\subsection{KL versus TF-IDF}
\label{sec:scoring}

The two rules trade the two forgetting metrics against each other.
Across architectures, KL keeps WikiText perplexity tighter than
TF-IDF, while TF-IDF retains more TriviaQA accuracy
(Table~\ref{tab:results}). The cleanest within-architecture contrast
is Additive +S: switching from TF-IDF to KL adds $+0.9$ pp on MedMCQA
but loses $2.2$ pp on TriviaQA. A plausible reason: KL more
aggressively picks the slots a batch reads heaviest, so the same
slots keep getting updated, while TF-IDF picks more broadly. If
those heavily-read slots also hold factual knowledge, KL's repeated
updates can disrupt it. The preferred
rule therefore depends on which forgetting metric matters more for
the deployment.

\section{Conclusion}
\label{sec:conclusion}

We re-implement SMF on \texttt{Qwen-2.5-0.5B-Instruct} and compare
against LoRA and full finetuning on MedMCQA. Two takeaways stand out.
Preserving the pretrained MLP path matters at this scale: replacement
memory, which discards the MLP, lies off both Pareto frontiers, while
the additive variant keeps the MLP available and lets a scaled memory
branch specialize. A plausible reason additive SMF reaches a
lower-forgetting region than LoRA or full finetuning is update
locality: LoRA adapters affect every input on every step, whereas
sparse memory updates only the small subset of value rows selected by
retrieval.

\textbf{Limitations.} We report this as a work in progress. Our
study uses one model (Qwen-2.5-0.5B), one target domain (MedMCQA),
and a 1{,}000-example slice, and we do not isolate update locality
from the dense retrofit stage. We probe forgetting through WikiText
perplexity and TriviaQA only, so capabilities such as instruction
following, code, and multi-step reasoning may degrade differently.
Pareto orderings may shift with larger models, longer training, or
different domains. The sparse-vs-LoRA contrast likely depends on how
broadly LoRA adapters are applied, and the KL-vs-TF-IDF tradeoff
depends on the background corpus (here OpenAssistant). Future work
could also explore domain-specific initialization of the memory
layers using more aligned corpora (e.g., PubMed for medical QA).

\bibliography{example_paper}
\bibliographystyle{icml2026}

\newpage
\appendix
\onecolumn
\section{Hyperparameters}
\label{app:hyper}

All training uses AdamW with a cosine learning-rate schedule, 100
warmup steps, max sequence length 1024, and gradient clipping at
$\Vert g \Vert = 1.0$. Per-method hyperparameters are listed in
Table~\ref{tab:hyper}.

\begin{table}[h]
  \caption{Per-method hyperparameters. ``Trained params'' counts the
  parameters that receive gradient at any point during training. For
  sparse methods the per-step trained-row count is much smaller (the
  top-$T = 512$ rows per layer).}
  \label{tab:hyper}
  \begin{center}
    \begin{tabular}{lcccc}
      \toprule
      Method & Trained params & LR & Epochs & Batch (eff.) \\
      \midrule
      Replacement sparse (values only)  & ${\sim}44$M & $5\!\times\!10^{-4}$ & 3 & 16 \\
      Additive sparse (values only)     & ${\sim}44$M & $5\!\times\!10^{-4}$ & 3 & 16 \\
      Additive sparse +S                & ${\sim}44$M $+\,3$ scalars & $5\!\times\!10^{-4}$ & 3 & 16 \\
      LoRA ($r=16$, $\alpha=32$)        & ${\sim}9$M  & $2\!\times\!10^{-4}$ & 3 & 16 \\
      Full finetune                     & ${\sim}494$M & $5\!\times\!10^{-5}$ & 3 & 16 \\
      \midrule
      Dense retrofit (replacement)      & ${\sim}52$M & $5\!\times\!10^{-4}$ & 2 & 16 \\
      Dense retrofit (additive)         & ${\sim}52$M $+\,3$ scalars & $5\!\times\!10^{-4}$ & 2 & 16 \\
      \bottomrule
    \end{tabular}
  \end{center}
\end{table}

Memory hyperparameters: $n_k = 128$ (so $M = 16{,}384$ slots per layer
of $d = 896$ each), 4 query heads, top-$k = 16$ neighbors per token at
inference, key dimension 256. Sparse-update top-$T = 512$ rows per
layer. Background statistics for slot scoring are collected from 2{,}000
batches of OpenAssistant with batch size 1.

\section{Update Locality}
\label{app:overhead}

The sparse methods differ from LoRA and full finetuning not only in
which parameters are trainable over the whole run, but also in how many
parameters receive gradient on a single step. Each memory layer has
$16{,}384$ value rows, and sparse task training selects $T=512$ rows
per layer per batch. Thus, during Stage 2, at most $3 \times 512$
memory rows receive value gradients in a step, or $3.125\%$ of the
value rows in each inserted memory. In parameter count, this is about
$1.38$M value parameters per step, compared with roughly $44$M memory
value parameters that may be updated over the full sparse run, roughly
$9$M trainable LoRA parameters, and roughly $494$M full-finetuned
parameters. This distinction is central to the SMF hypothesis:
localizing each update may reduce interference even when the total
adaptation substrate is large.

Table~\ref{tab:overhead} separates three quantities that are easy to
conflate: parameters stored for the adaptation, net inference-time model
size, and parameters updated on each sparse task-training step. The
three selected Qwen MLPs contain roughly $39.2$M parameters
($3$ layers $\times$ gate/up/down projections with $d_{\mathrm{model}}
=896$ and $d_{\mathrm{ff}}=4864$). The three inserted memory modules
contain roughly $52$M parameters, of which about $44$M are value rows.
Thus replacement memory has modest net parameter overhead because it
removes the selected MLPs, whereas additive memory keeps the original
MLPs and adds the memory modules on top.

\begin{table}[h]
  \caption{Storage, inference-size, and per-step update comparison.
  Counts are approximate for Qwen-2.5-0.5B with memory at layers
  $\{6,12,18\}$. ``Updated per Stage-2 step'' counts parameters that
  receive gradient during sparse MedMCQA task training, not during the
  dense retrofit stage.}
  \label{tab:overhead}
  \vskip 0.05in
  \begin{center}
    \begin{small}
      \begin{tabular}{lccc}
        \toprule
        Method & Adaptation params stored & Net inference-size change & Updated per Stage-2 step \\
        \midrule
        Full finetune & 0 extra & $0$ & ${\sim}494$M \\
        LoRA & ${\sim}9$M adapters & $0$ if merged; ${\sim}{+}9$M unmerged & ${\sim}9$M \\
        Replacement memory & ${\sim}52$M memory & ${\sim}{+}13$M (${\sim}{+}2.6\%$) & ${\sim}1.38$M value params \\
        Additive memory & ${\sim}52$M memory & ${\sim}{+}52$M (${\sim}{+}10.5\%$) & ${\sim}1.38$M value params \\
        Additive memory +S & ${\sim}52$M memory $+3$ scalars & ${\sim}{+}52$M (${\sim}{+}10.5\%$) & ${\sim}1.38$M value params $+3$ scalars \\
        \bottomrule
      \end{tabular}
    \end{small}
  \end{center}
\end{table}

\section{Eval Details}
\label{app:eval}

We evaluate on 1{,}000-example slices of each test set. MedMCQA
multiple-choice scoring uses the mean per-token negative log-likelihood
of each candidate continuation given the prompt. This avoids the
length bias of the corresponding sum-log-likelihood scorer. WikiText
perplexity is computed by concatenating the wikitext-103-v1 test split,
tokenizing once, then summing token-level negative log-likelihood under
a sliding window of size 1024 with stride 512 (each token contributes
to the loss exactly once). TriviaQA uses the rc.nocontext validation
split with greedy decoding for 32 new tokens. An example is counted
as correct if any answer alias appears as a substring of the
normalized prediction.

\section{LLM Usage}
\label{app:llm}

We used LLMs (ChatGPT and Claude) to assist with writing edits,
\LaTeX{} formatting, and code. All technical content,
experiments, and claims were produced and verified by the authors.

\end{document}